\documentclass[runningheads]{llncs}
\usepackage{graphicx}
\usepackage{bm}
\usepackage{subfigure}
\usepackage{algorithm}  
\usepackage{algorithmicx}  
\usepackage{algpseudocode}  
\usepackage{amsmath}
\usepackage[T1]{fontenc}
\usepackage{mathrsfs}
\usepackage{enumitem}

\begin{document}

\title{Semantic Aware Attention Based Deep Object Co-segmentation} 
\titlerunning{Semantic Aware Attention Based Deep Object Co-segmentation} 


\author{Hong Chen \and
Yifei Huang \and
Hideki Nakayama}
%

\authorrunning{Hong. et al.} 


\institute{The University of Tokyo\\
\email{chen@nlab.ci.i.u-tokyo.ac.jp}\\
\email{hyf@iis.u-tokyo.ac.jp}\\
\email{nakayama@ci.i.u-tokyo.ac.jp}}

\maketitle

\begin{abstract}
Object co-segmentation is the task of segmenting the same objects from multiple images.
In this paper, we propose the Attention Based Object Co-Segmentation for object co-segmentation that utilize a novel attention mechanism in the bottleneck layer of deep neural network for the selection of semantically related features. 
Furthermore, we take the benefit of attention learner and propose an algorithm to segment multi-input images in linear time complexity.
Experiment results demonstrate that our model achieves state of the art performance on multiple datasets, with a significant reduction of computational time.

\keywords{Co-segmentation  \and Attention \and Deep learning.}
\end{abstract}
\section{Introduction}

\begin{figure}[h!]
	\centering
    \includegraphics[scale=0.2]{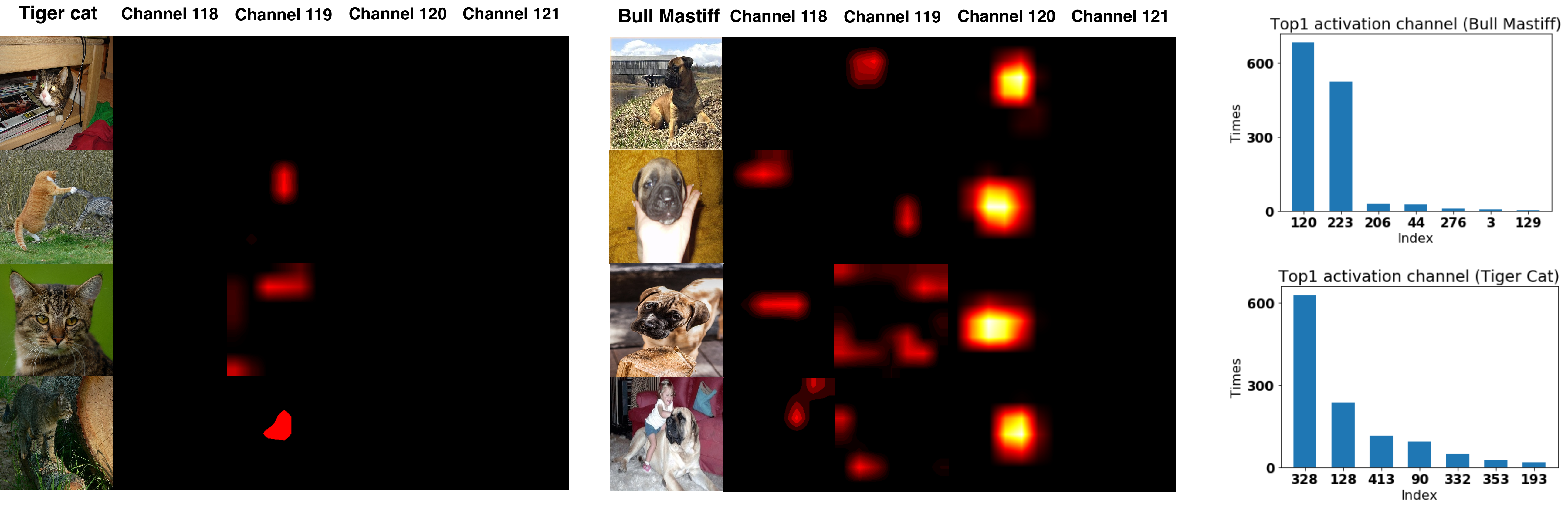}
	\caption{Visualization activation of channels 118 - 121 (\textit{conv5\_3 layer of VGG16}) with Grad-Cam\cite{gradcam} for two different classes: (a) Bull Mastiff (b) Tiger Cat. Different classes correspond to different layer activations. We leverage this property and propose our semantic aware attention based object co-segmentation model, by using an attention learner to find relevant channels to enhance and irrelevant channel to suppress.}
	\label{fig:cam1}
\end{figure}

Image segmentation is one of a fundamental computer vision problem which aims to segment images into semantically different regions. Recently, remarkable success have been made based on the rapid development of deep learning~\cite{semantic1,semantic2,semantic3,semantic4,semantic5}. First proposed by Rother et al.~\cite{objectcoseg1}, \textit{object co-segmentation} which aims at extracting similar objects from multiple inputs, utilizes joint information from two images and achieves higher accuracy compared to segmenting the objects independently. This can be used in various applications like image retrieval~\cite{objectcoseg1} and object discovery~\cite{objectcoseg4}.


While considerable attention have been paid to object segmentation, there are limited previous works focusing on object co-segmentation~\cite{objectcoseg2,objectcoseg3,objectcoseg4,objectcoseg1,objectcoseg5,objectcoseg6,objectcoseg8,objectcoseg9,objectcoseg10,objectcoseg11,objectcoseg12}. 
Intuitively thinking, the advantage of co-segmentation against segmentation is to jointly utilize information from both images so as to perform better segmentation. In particular these information include 1)appearance similarity, 2)background similarity and 3)semantic similarity. There are previous works that leverage 1) and 2) \cite{objectcoseg5,objectcoseg6}, but these are not general since appearance or background is not always similar. Recently, a deep learning based method \cite{objectcoseg2} focused on semantic similarity, they can co-segment objects in the same semantic class even with different appearance and background, and outperformed other conventional methods by a large margin. They use a correlation layer \cite{flownet} to compute localized correlations between semantic features of two input images, then predict the mask of common objects. However, since the correlation is computed in a pair-wise manner, their method is hard to extend to co-segmentation of more than two images. If the number of images in the group is large, the time complexity of co-segmentation with multiple (i.e, more than three) images will increase drastically when considering all different pairs in the group.

In this work we aim to co-segment the objects of a same semantic class in multiple input images, even with different appearance and background. We also aim to enable instant group co-segmentation, without suffering extra time complexity by pair-wisely considering all possible image pairs.

In a deep neural network, higher abstract semantic information is encoded in deeper layers, and different channels correspond to different semantic meanings \cite{visualize}. Figure \ref{fig:cam1} is a visualization of the channel activation of the \textit{conv5\_3} layer of VGG16 to different input images. We can see that strong activations are observed in channel 120 and 223 with respect to class \textit{Bull Mastiff}, while the same channel only have very weak activations with respect to class \textit{Tiger Cat}. Motivated by this observation, we argue that by applying attention in deep features (i.e, emphasizing channels whose activation is strong in both images' features and suppressing the other channels), semantic information can be selected and enhanced, thus co-segmentation can be performed. According to this disentangled property, when dealing with multiple inputs, we can regard attentions as semantic selectors which can be applied globally instead of taking care of intra semantic relationship pair-wisely. Also, attention learner can be effectively implemented as fully connected layers and average pooling layers, which makes it faster than correlation layer used in \cite{objectcoseg2} in one forward operation.


In this paper, we propose a novel attention based co-segmentation model that leverages attention in bottleneck layers of the deep neural network. The proposed model is mainly composed of three modules: an encoder, a semantic attention learner, and a decoder. The encoder encodes the images into highly abstract features, in this work we take the convolutional layers of VGG16 as the encoder. The semantic attention learner takes the encoded feature and learns to pay attention to the co-existing objects. We propose three mechanisms for this attention learner and will describe them in detail in Section \ref{sec:model}. The decoder then uses the attended deep features to output the final segmentation mask. 


We summarize the main contributions of this work as follows:
\begin{itemize}
\item   We propose a simple yet efficient deep learning architecture for object co-segmentation: Attention Based Deep Object Co-segmentation model. In our model, we use a semantic attention learner to spotlight feature channels that have high activation in all input images and suppress other irrelevant feature channels. To the best of our knowledge, this is the first work that leverages attention mechanism for deep object co-segmentation.  
\item   Compared with previous works that perform co-segmentation in quadratic time complexity, our proposed model can do co-segmentation of multiple images in linear time complexity.
\item   Our model achieves state of the art performance in multiple object co-segmentation datasets and is able to generalize to unseen objects absent from the training dataset.
\end{itemize}

\section{Related work}

\subsubsection{Object Co-segmentation}

The term Object Co-segmentation that segment
"object" instead of "stuff" was first proposed by Rother et al. \cite{objectcoseg1} in 2011. Rubinstein et al. \cite{objectcoseg4} captured the sparsity and visual variability of the common object over the entire database using dense correspondences between images to avoid noise while finding saliency. Utilizing the clustering method, Joulin et al. \cite{objectcoseg6} pointed out that discriminative clustering is well adapted to the co-segmentation problem and they extended its formulation to fit the co-segmentation task.

With the assistance of deep learning, Mukherjee et al.~\cite{objectcoseg3} generated object proposals from two images and turned them into vectors by a Siamese network. During the training, they built an Annoy (Approximate Nearest Neighbor) Library to measure their Euclidean distance or Cosine distance between two vectors. Recently, DOCS~\cite{objectcoseg2} turned PASCAL VOC dataset into a co-segmentation dataset, producing more paired data. They applied a correlation layer \cite{flownet} to find out similar features. They proposed group co-segmentation to test on several datasets and demonstrated that their result has achieved state-of-the-art. However, their pairwise scheme makes the testing cost a lot of computation time when performing group co-segmentation, so testing on the whole Internet dataset becomes computationally intractable. They only tested on the subset of it. In this paper, our method can run in linear time and in the meantime, achieve state of the art performance. Our proposed model has a much simpler structure, yet can still get better performance.

\subsubsection{Pixel Objectness}
Pixel Objectness was first proposed in \cite{pixelobjectness}. Their research revealed that the feature map from the model pretrained by Imagenet \cite{imagenet} could be used in the task called \textit{object discovery}. Similar with our model, they extended a decoder after the last convolutional layer of VGG16 to produce the segmentation mask. Though they only trained the model on PASCAL VOC Dataset, their model can segment other objects even not existed in the training dataset. 

Our model extended their work with a novel attention learner. By this means, we can not only perform image co-segmentation task, but also enhance the model's ability for object discovery.

\subsubsection{Attention}

As far as we know, visual attention was first proposed in \cite{attention1}. Since then many attention based models have been used in many computer vision tasks such as VQA \cite{attention9,attention10,attention13,attention14} and image captioning \cite{attention8,attention2,attention1,attention12}. Recently, attention models have been widely used in many other research domains \cite{attention3,attention4,attention5,attention6,attention7,huang2017temporal}. Attention can be considered as laying weights on channels of feature map to enhance some semantic information and, at the same time, remove other unwanted semantic information. 

In our paper, we utilize the channel-wise attention model as semantic selectors, since specific channels contribute to the specific semantic class. In this object co-segmentation task, we generate channel-wise attention from one image to decide which semantic information should be removed in the other. To the best of our knowledge, this is the first work to use channel-wise attention model in the task of object co-segmentation.

\section{Model}\label{sec:model}
\begin{figure}[h!]
	\centering
    \subfigure[CA]{\label{fig:model:a}\includegraphics[scale=0.4]{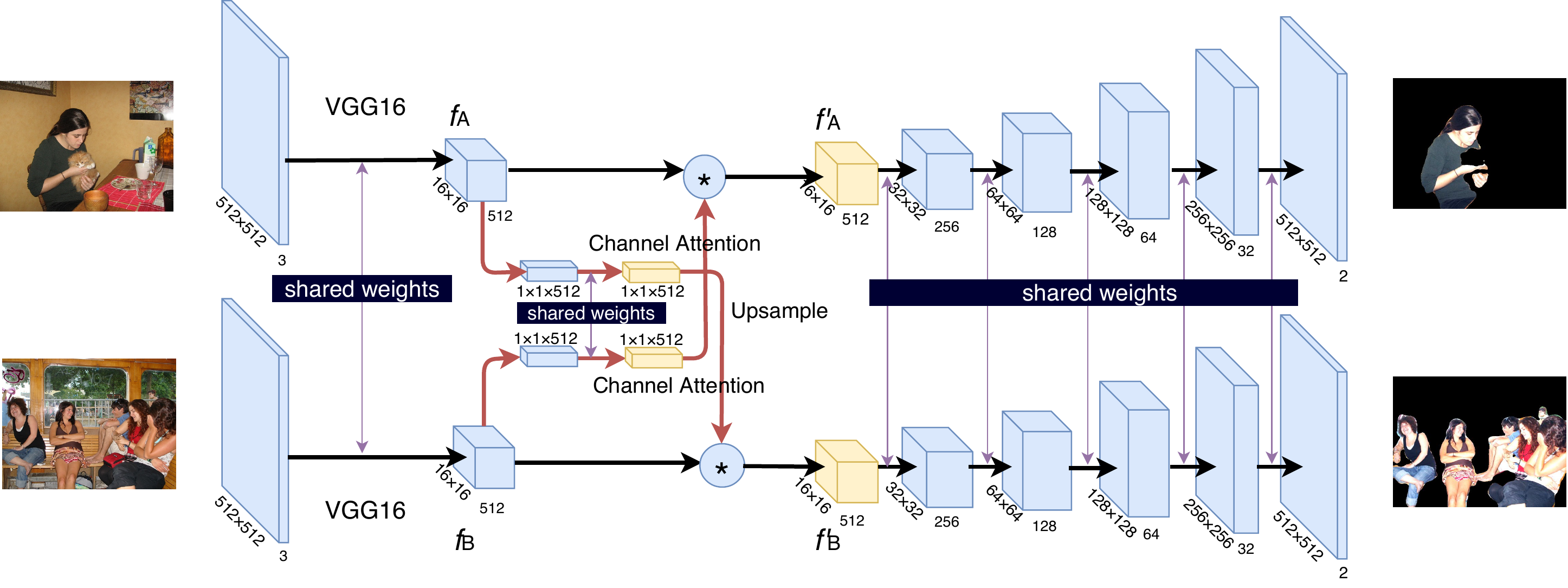}}
    $\vcenter{\hbox{\subfigure[FCA]{\label{fig:model:b}\includegraphics[scale=0.4]{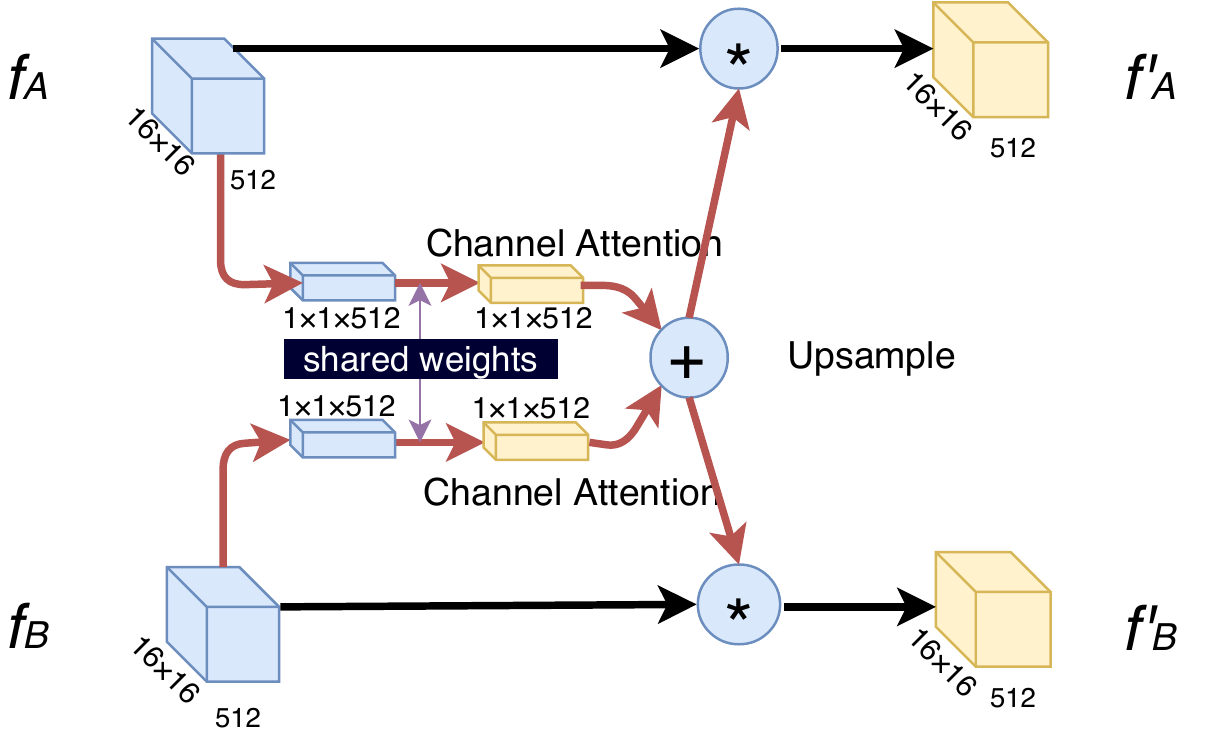}}}}$
    $\vcenter{\hbox{\subfigure[CSA]{\label{fig:model:c}\includegraphics[scale=0.4]{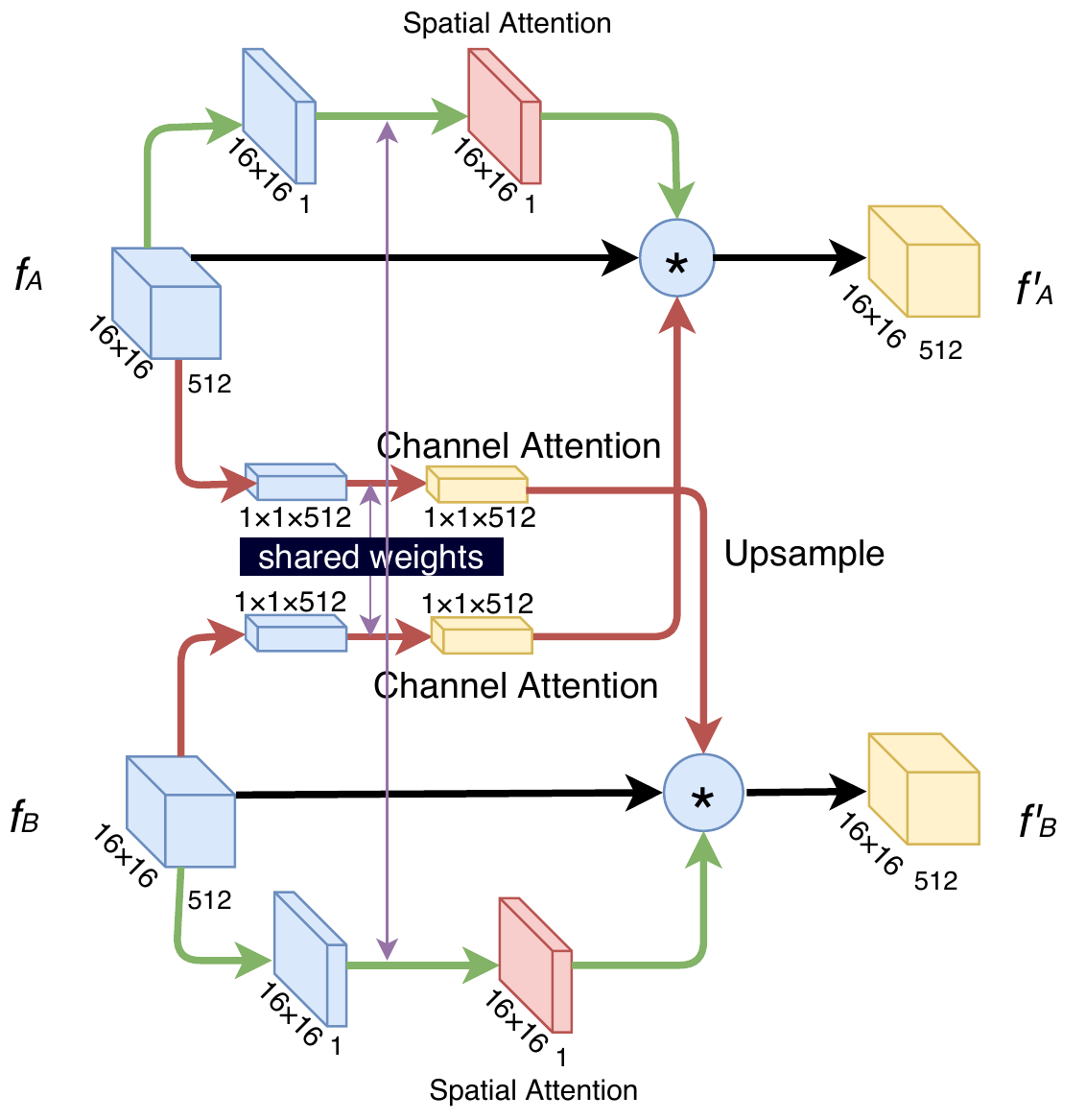}}}}$
	\caption{Model overview. Our model contains three parts: Siamese encoder, attention learner and Siamese decoder. Encoder uses general VGG16, and decoder uses the up-sampling method. In the middle part, we propose three different architectures of attention learner: CA, FCA and CSA. (a) Channel-wise Attention (CA): Generating the channel-wise attention from each image separately and apply it on the other feature map. (b) Fused Channel-wise Attention (FCA): Combining both attentions and generate a new attention for both images. (c) Channel Spatial Attention (CSA): Generating the spatial-wise attention as well as the channel-wise attention.}
	\label{fig:model}
\end{figure}

Fig.~\ref{fig:model} presents the overview of our model. For simplicity, we demonstrate our model using two inputs, which is the typical case of co-segmentation. We will show in Section \ref{sec:group} that our model can be extended to multiple inputs when testing. Our model is composed of an encoder, a semantic attention learner, and a decoder. The encoder is identical with the convolutional layers of VGG16, thus the output from the encoder is a 512-channel feature map. The feature maps are forwarded into the semantic attention learner.
Here we propose three different architectures for the semantic attention learner: channel-wise attention model (CA), fused-channel-wise attention model (FCA) and channel-spatial-wise attention model (CSA).
Processed by semantic attention learner, we obtain the attended feature map and forward it to the decoder. 
We adopt the up-sampling method for the decoder and add a dropout layer after each layer to avoid overfitting. 
For the last layer, we use a convolution layer to output two channels, representing the foreground and the background respectively.
To be emphasized, unlike \cite{objectcoseg2}, we do not concatenate the original feature with the processed feature since the innovative feature tends to segment every "object-like stuff" and ignore the semantic selector, which is contrary to the goal of this task.

\subsection{Channel Wise Attention(CA)}
It is known that each channel contains different semantic information with respect to different semantic classes~\cite{visualize} as mentioned in Fig ~\ref{fig:cam1}. 
We construct the Channel-wise attention (CA) to help us enhance the semantic information we need and suppress the remaining. To be specific, we first do global average pooling for the output of the encoder $f{_A}$ and $f{_B}$ and forward to a learnable fully connected (FC) layer to get two weight vectors $\bm{\alpha}_A, \bm{\alpha}_B \in R^{512}$. $\bm{\alpha}_A, \bm{\alpha}_B$ contain semantic information of two inputs: if the $i$-th index of $\bm{\alpha}_A$ is large, it indicates the $i-th$ channel has high activation, and thus the image contains the semantic represented by channel $i$. By performing channel-wise multiplication with $\bm{\alpha}_A$ and $f{_A}$, $\bm{\alpha}_B$ and $f{_B}$, we can get the attended feature maps $f'{_A}$ and $f'{_B}$. Here $f{_A}$ receives the attention weights computed from $f{_B}$, so if $f{_A}$ has high activation in channel $i$ while $f{_B}$ contains no such semantic ($\bm{\alpha}_{B,i}$ is small), this channel will be suppressed. On the other hand, if $f{_A}$ does not have high activation in channel $j$ while the $j$-th index of $\bm{\alpha}_B$ is large, channel $j$ will not be activated since this channel activation is initially small. Same will happen to $f{_B}$. As a consequence, by applying this channel wise attention mechanism, only the channels with high activations in both inputs are preserved and enhanced relatively. Other channels which do not contain semantics in both inputs will be suppressed. We also show the property of the channel wise disentanglement in our supplementary material.
We present details in Equation (1) - (4). In all of the equations below, * represents matrix multiplication, and $\odot$ represents element-wise multiplication with broadcasting.
\begin{flalign}
      \bm{\alpha}_A&=\sigma(W{^T}*AvgPool_{channel}(f{_A})+b) \\
      \bm{\alpha}_B&=\sigma(W{^T}*AvgPool_{channel}(f{_B})+b) \\
      f'{_A}&=\ \bm{\alpha}_B\odot f{_A} \\
      f'{_B}&=\ \bm{\alpha}_A\odot f{_B} 
  \label{equ:attention1}
\end{flalign}
Note that here we use a sigmoid function instead of the generally used softmax function because our approach is to use attentions for retaining or removing semantic information. If two objects with different semantics both appear in both images, according to our assumption, the weights of related channels of the objects appear in the other image need to be close to 1 in order to maintain their information. However, if we use a softmax layer, both related and irrelevant channel weights may become less than 0.5, which will affect the performance of attention co-segmentation. 


\subsection{Fused Channel Wise Attention(FCA)}
Since we aim to find the same semantic information in both inputs, the output attention weights from the semantic attention learner for both input images should be the same. Therefore, we propose another way of finding common semantic information from $\bm{\alpha}_A$ and $\bm{\alpha}_B$, by using one FC layer to fuse the two attention weights. Fig.~\ref{fig:model:b} shows the architecture of our fused channel-wise attention.
We take the same procedure from CA to generate $\bm{\alpha}_A$ and $\bm{\alpha}_B$.
$\mathcal{A}_C$ is the combined results of both attentions:
\begin{flalign}
      \bm{\alpha}_C&=\sigma(W{^T}*(\bm{\alpha}_A+\bm{\alpha}_B)+b) \\ 
      f'_A&=\bm{\alpha}_C\odot f_A \\
      f'_B&=\bm{\alpha}_C\odot f_B 
  \label{equ:attention2}
\end{flalign}

\subsection{Channel Spatial Attention(CSA)}
While Global Average Pooling can extract global information from the feature map, spatial information will be lost due to this pooling operation. To further improve CA, inspired by \cite{attention2}, we propose to lay some spatial information for further improving its segmentation performance. We can find that our spatial attention produces reasonable heat map on objects in different pictures. We show visualizations of spatial wise attention in supplementary material.

For this channel spatial attention architecture, we first generate channel wise attention $\bm{\alpha}^c_A, \bm{\alpha}^c_B$ using the same fashion as CA. For generating the spatial-wise attention, we calculate mean value of each spatial location across all channels to generate spatial attention maps $\bm{\alpha}^s_A, \bm{\alpha}^s_B \in R^{W\times H}$:
\begin{flalign}
      \bm{\alpha}^c_A&=\sigma(W{^T}*AvgPool_{channel}(f_A)+b) \\
      \bm{\alpha}^c_B&=\sigma(W{^T}*AvgPool_{channel}(f_B)+b) \\
      \bm{\alpha}^s_A&=\sigma(AvgPool_{spatial}(f_A))\\
      \bm{\alpha}^s_B&=\sigma(AvgPool_{spatial}(f_B))\\
      f'_A&=\bm{\alpha}^c_B\odot \bm{\alpha}^s_A\odot f_A \\
      f'_B&=\bm{\alpha}^c_A\odot \bm{\alpha}^s_B\odot f_B  
  \label{equ:attention3}
\end{flalign}

\subsection{Instant Group Co-segmentation} \label{sec:group}
With group co-segmentation proposed in \cite{objectcoseg2}, we have to match all possible image pairs separately. Co-segmenting $n$ images needs quadratic $\big(O(n^2)\big)$ computation time. Thus, it becomes computationally intractable to test on a large scale dataset. Moreover, because of their fixed structure, the existing deep learning models shows weaknesses when co-segmenting more than two images at the same time.
For example, we can hardly use \cite{objectcoseg2} to discover and segment the most frequently appearing object among several (more than 2) images, since it is hard to measure frequency due to the lack of global semantical understanding in correlation modules.

To address this problem, thanks to our attention mechanism, we present an approach to accomplish co-segmentation in a single shot by controlling their generated attentions. Without loss of generality, we use CA (the first proposed model) to demonstrate the procedure.

\begin{algorithm}[h!]  
  \caption{Instant Group Co-segmentation}  
  \label{alg::instant}  
\textbf{Input} 
  \hspace*{\algorithmicindent}      $N$: Numbers of inputs.    \hspace*{\algorithmicindent}$I$: Images $I{_1}$,$I{_2}$,$I{_3}$\dots $I{_N}$.\\
\textbf{Output}
   \hspace*{\algorithmicindent}    $S$: Segment mask $S{_1}$,$S{_2}$,$S{_3}$\dots $S{_N}$.
  \begin{algorithmic}[1]  
    \State Split the model into AttentionGeneration() and Segmentation().
    \For {$i={1,2,3...N}$} 
      \State compute attention $\bm{\alpha_i} = AttentionGeneration(I_i)$;
    \EndFor
    \State Calculate Group Average Attention from all $\bm{\alpha_i}$. $\bm{\alpha_{avg}} = mean (\bm{\alpha})$
     \For {$i={1,2,3...N}$}
      \State $S{_i}$ = Segmentation($I_i$,$\bm{\alpha_{avg}}$)  
     \EndFor
  \end{algorithmic}  
\end{algorithm} 

Although our model is trained end to end by pairs of input images, when testing, our model can be seen as a composition of two parts: an attention-generating module and a segmentation module. When forwarding images $\{I_1, \cdots ,I_N\}$ into our model, each image $I_k$ will generate an attention weight $\bm{\alpha}_k$. Each $\bm{\alpha}_k$ represents the disentangled semantic information of each $I_k$. To get the common semantic information for all the input images, we take the averaged attention weight $\bm{\overline{\alpha}}=\sum_{i=1}^k \bm{\alpha}_k$ and use $\bm{\overline{\alpha}}$ for attending on each image features. We name this procedure \textbf{Group Average Attention} and show in detail in Algorithm \ref{alg::instant}. Thus, we reduce time complexity into linear time $\big(O(n)\big)$. Based on specific task and dataset, the global average attention can also be changed to global minimum attention, by changing the averaging operation to minimum operation of all $\bm{\alpha}_k$ along all dimensions. By this means we can strictly co-segment objects that appear in all input images.
\subsection{Training and Implementation Details}
We use PyTorch\cite{pytorch} library to implement our model. For the training, we rescale every image into spatial size 512*512. 
As for the decoder we adopt the architecture: upsample$\rightarrow$ conv (512,256)$\rightarrow$ upsample$\rightarrow$ conv (256,128)$\rightarrow$ upsample$\rightarrow$ conv (128,64)$\rightarrow$ upsample$\rightarrow$ conv (64,32)$\rightarrow$ upsample$\rightarrow$ conv (32,2), in which conv (a,b) indicates a convolution layer whose input channel is a and output channel is b. For all convolution layers we use kernel size 3 and stride 1, followed by a Rectified Linear Unit (ReLU) layer and Batch Normalize Layer\cite{batchnorm}. We compute loss using cross entropy loss, and back propagate gradients to all layers. We use Adam\cite{adam} optimizer with learning rate 1e-5 for optimization.


\section{Experiments}
\subsection{Datasets}
For training, we use the same dataset as \cite{objectcoseg2}, where the image pairs are extracted from the PASCAL VOC 2012 training sets. The total number of pairs is over 160k. 

For validation and testing our model, we randomly separate the PASCAL VOC2012 validation set into 724 validation images and 725 test images like \cite{objectcoseg2}, by pairing the images we get 46973 and 37700 pairs respectively.

We also test our model using other datasets commonly used in object co-segmentation. They include:
\begin{itemize}
\item MSRC~\cite{MSRC1,MSRC2} sub-dataset. In this dataset there are 7 classes: bird, car, cat, cow, dog, plane, sheep. Each class contains 10 images.
\item Internet~\cite{objectcoseg4} sub-dataset, each of the 3 classes: car, horse and airplane contains 100 images.
\item Internet~\cite{objectcoseg4} dataset. The three classes car, horse and airplane contain 1306, 879 and 561 images correspondingly. Note that the MSRC dataset and two Internet dataset all contains classes within the training data. Here we refer to these object classes as \textit{Seen Objects}.
\item ICoseg~\cite{icoseg} sub-dataset. This dataset contains 8 classes, each with a different number of images. Different from the previous datasets, the class in ICoseg dataset is different from the training dataset. We adopt this dataset to test the generalizability of our model. We refer to the objects in ICoseg dataset as \textit{Unseen Objects}.
\end{itemize}

\subsection{Baselines}
For result comparison, we use the following baselines: \cite{objectcoseg1} is a conventional method which first extracts the features from image pairs and then trains a Random Forest Regressor based on these features. \cite{objectcoseg4,objectcoseg11,objectcoseg12} do co-segmentation based on saliency. \cite{objectcoseg6,objectcoseg8} utilize clustering method to find the similarity of the image. \cite{objectcoseg10} uses conditional random fields to find the relationship between pixels.
\cite{objectcoseg9} connects the superpixel nodes located in the image boundaries of the entire image group, then infers via the proposed GO-FMR algorithm.
\cite{objectcoseg5} first induces affinities between image pairs and then co-segments objects from co-occurring regions. They further improve their results with Consensus Scoring.
Other than the pre deep learning conventional methods,
\cite{objectcoseg3,objectcoseg2} both utilized deep learning method for object co-segmentation.
\cite{objectcoseg3} found the objects with object proposals, encoded them into feature vectors with VGG16, then they compare the similarity of the feature vectors and decided which to segment.
Recently, \cite{objectcoseg2} proposed an end-to-end deep learning model with VGG16, a correlation layer and a decoder, which made great progress and achieved state of the art in this area. 

Pixel Objectness\cite{pixelobjectness} is the state of the art method of object discovery, and it is capable of segmenting all the existing objects with a single input image. The network architecture of Pixel Objectness is a VGG16 encoder directly followed by a decoder. So in the view of network architecture, our model can be viewed as adding attention learner in the middle of Pixel Objectness. So since the task is different, it is unfair to compare their method with co-segmentation methods quantitatively. We, therefore, demonstrate qualitative results to show that our model can inherit (and even perform better than) the object discovery ability of Pixel Objectness.

\subsection{Results and visualization of co-segmentation}\label{sec:resultCoseg}

\setlength{\tabcolsep}{4pt}
\begin{table}[h!]
\begin{center}
\caption{
Results on MSRC sub dataset with pairwise co-segmentation.
}
\label{table:msrc}
\begin{tabular}{ccccccccc}
\hline\noalign{\smallskip}
 MSRC & \cite{objectcoseg1} & \cite{objectcoseg4} & \cite{objectcoseg5} &  \cite{objectcoseg3} & \cite{objectcoseg2} & CA(ours) & FCA(ours) &CSA(ours) \\
\noalign{\smallskip}
\hline
\noalign{\smallskip}
Precision  & 90.2 & 92.16 & 92.0 & 84 & 92.43 & \textbf{96.6}& 96.31& 95.26\\
Jaccard & 70.6 & 74.7 & 77 & 67 & \textbf{79.89} & 76.49& 76.94& 77.7\\
\hline
\end{tabular}
\end{center}
\end{table}
\setlength{\tabcolsep}{1.4pt}


We first show the performance comparisons of conventional pairwise co-segmentation. For PASCAL VOC Datasets, we test our models on 37700 images pairs. We gain the Jaccard accuracy 59.24\%, 59.41\% and 59.76\% for CA, FCA and CSA respectively. Table~\ref{table:msrc} and~\ref{table:internet} show quantitative results on different MSRC and Internet sub-dataset. We can see that our model is able to get state of the art performance when segmenting \textit{Seen Objects}. Among the three proposed attention mechanisms, CSA is best at the co-segmentation of \textit{Seen Objects}. 

\setlength{\tabcolsep}{4pt}
\begin{table}[h!]
\begin{center}
\caption{
Results comparison on Internet sub dataset with pairwise co-segmentation.
}
\label{table:internet}
\begin{tabular}{ccccc}
\hline\noalign{\smallskip}
Internet(Jaccard) & Cars & Horse & Airplane &  Average \\
\noalign{\smallskip}
\hline
\noalign{\smallskip}
\cite{objectcoseg6}  & 37.1 & 30.1 & 15.3 & 27.5 \\
\cite{objectcoseg4} & 64.4 & 51.6 & 55.8 & 57.3 \\
\cite{objectcoseg8} & 64.9 & 33.4 & 40.3 & 46.2 \\
\cite{objectcoseg9} & 66.8 & 58.1 & 56.3 & 60.4 \\
\cite{objectcoseg10} & 72.0 & 65.0 & 66.0 & 67.7 \\
\cite{objectcoseg2} & \textbf{82.7} & 64.6 & 63.5 & 70.3 \\
CA(ours) & 80.0& 67.3& \textbf{71.4}& 72.8\\
FCA(ours) & 76.9& 69.1& 65.9& 70.6\\
CSA(ours) & 79.9& \textbf{71.4}& 68.0& \textbf{73.1}\\
\hline
\end{tabular}
\end{center}
\end{table}
\setlength{\tabcolsep}{1.4pt}
Table~\ref{table:icoseg} shows quantitative result of \textit{Unseen Objects} from the ICoseg sub dataset. We can see that our model (especially FCA) outperforms all baseline methods. In particular, we get 1.8\% performance gain compared with \cite{objectcoseg2} which also uses deep learning based method.

\setlength{\tabcolsep}{4pt}
\begin{table}[h!]
\begin{center}
\caption{
Results on ICoseg sub-dataset with pairwise co-segmentation.
}
\label{table:icoseg}
\begin{tabular}{cccccccccc}
\hline\noalign{\smallskip}
iCoseg(Jaccard) & \cite{objectcoseg4} & \cite{objectcoseg11} & \cite{objectcoseg5} &  \cite{objectcoseg12} & \cite{objectcoseg2} & CA(ours) & FCA(ours) &CSA(ours) \\
\noalign{\smallskip}
\hline
\noalign{\smallskip}
bear2 &65.3&70.1&72.0&67.5&88.3&90.2&88.3&\textbf{90.6}\\
brownbear &73.6&66.2&92.0&72.5&92.0&92.9&91.5&\textbf{93.0}\\
cheetah &69.7&\textbf{75.4}&67.0&78.0&68.8&65.7&71.3&51.4\\
elephant &68.8&73.5&67.0&79.9&84.6&82.5&84.4&\textbf{85.1}\\
helicopter &80.3&76.6&\textbf{82.0}&80.0&79.0&74.2&76.5&77.1\\
hotballoon &65.7&76.3&88.0&80.2&91.7&92.9&\textbf{94.0}&93.7\\
panda1 &75.9&80.6&70.0&72.2&82.6&\textbf{92.2}&91.8&92.5\\
panda2 &62.5&71.8&55.0&61.4&86.7&\textbf{91.1}&90.3&89.1\\
\hline
\noalign{\smallskip}
average &70.2&73.8&78.2&70.4&84.2&85.2&\textbf{86.0}&84.0\\
\noalign{\smallskip}
\hline
\end{tabular}
\end{center}
\end{table}
\setlength{\tabcolsep}{1.4pt}

\begin{figure}[h!]
	\centering
     \subfigure[Seen Object(i.e, the class in the dataset)]
    {\label{fig:demo:seen}\includegraphics[scale=0.36]{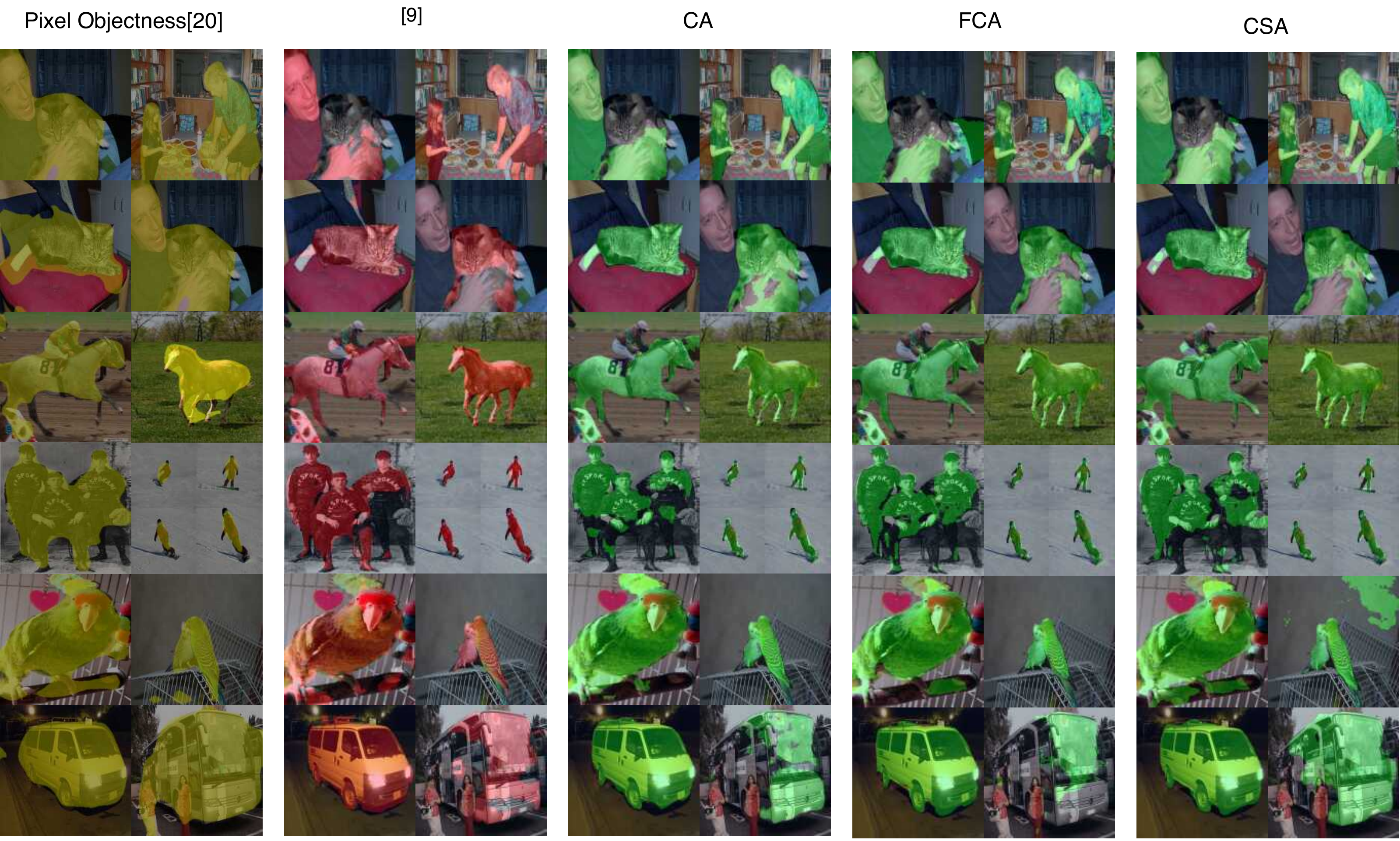}}
    \subfigure[Unseen Object(i.e, the class beyond the dataset)]
    {\label{fig:demo:unseen}\includegraphics[scale=0.36]{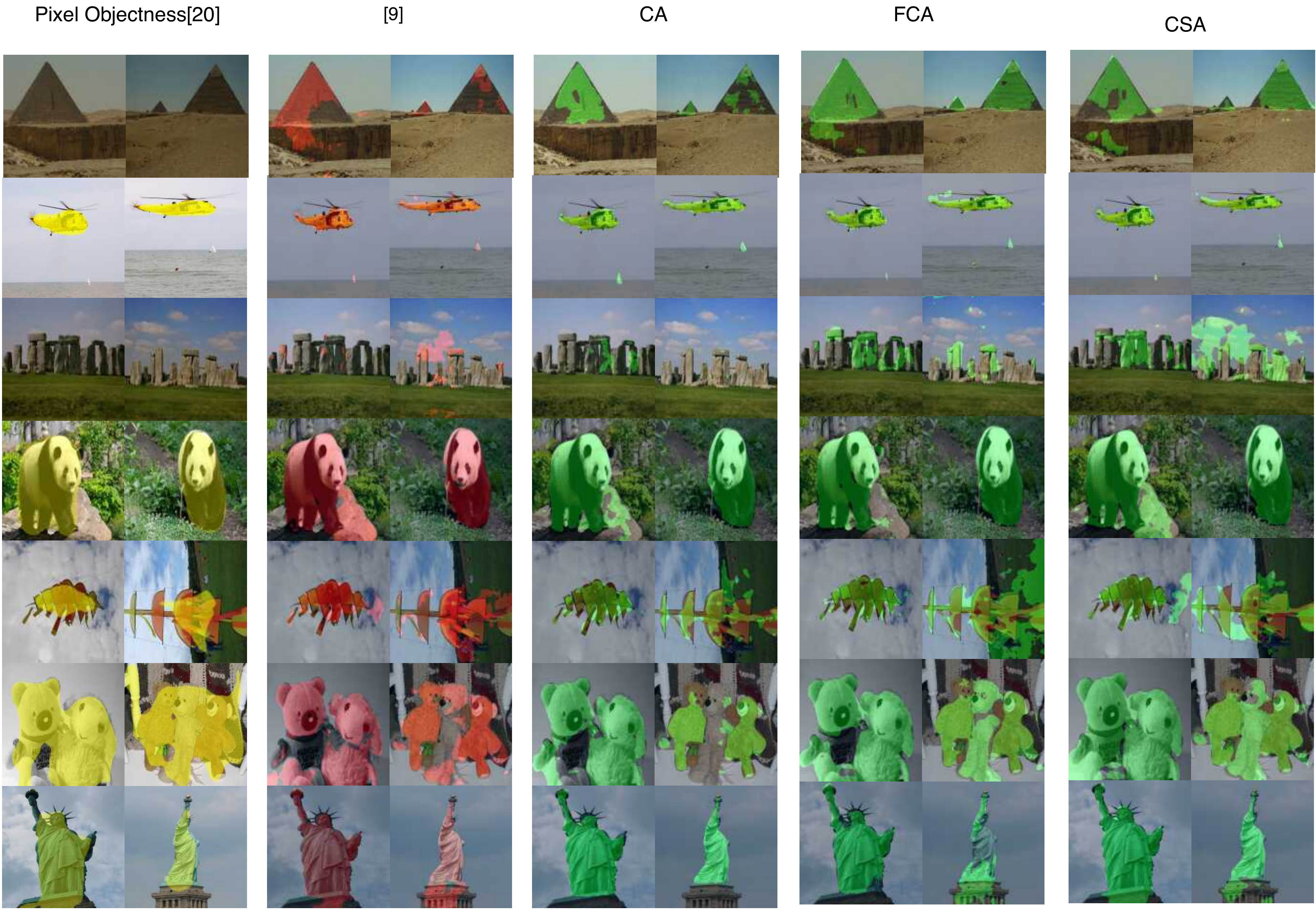}}

	\caption{Qualitative co-segmentation results. Each row indicate the input image pairs to different methods. Red masks are the output by \cite{objectcoseg2}, yellow by Pixel Objectness \cite{pixelobjectness}, and the output of our proposed methods are shown in green masks.}
	\label{fig:demo}
\end{figure}

In Fig.~\ref{fig:demo}, we visualize some co-segmentation results from three models. These demonstration images are selected from all of the three datasets containing \textit{Seen Objects}: MSRC, Internet, PASCAL VOC and MSCOCO dataset. We can see that Pixel Objectness \cite{pixelobjectness} tends to over segment since it outputs the segmentation of all possible objects. We can see that our method performs comparably well as the state of the art deep learning based method \cite{objectcoseg2}. We will further demonstrate that our method has much less time complexity compared with \cite{objectcoseg2} in Section \ref{sec:instalresult}.

In Fig.~\ref{fig:demo:unseen}, the comparison between our method and Pixel Objectness \cite{pixelobjectness} demonstrates that our approach has a stronger ability in object discovery, thanks to the attention learner that can help to reduce semantic noise and enhance some semantic information in the feature map. For example, from the first row in Fig. \ref{fig:demo:unseen}, with Pixel Objectness, no object could be detected but assisted with reference attention from another image, our model can detect and segment the pyramid precisely. Also, we can see that our method achieves better performance in co-segmentation task compared with \cite{pixelobjectness} and \cite{objectcoseg2}. FCA is the best architecture for co-segmentation of \textit{Unseen Objects}. 
According to the architecture, CSA remains most information from the original images because of the spatial attention. However, it is unclear that spatial attention also has the disentangle properties which may lead to miss-segment in the unseen objects.
On the other hand, FCA aims at finding the common attention between two inputs, so noises from two generated attention will be suppressed.

\subsection{Instant group co-segmentation results}\label{sec:instalresult}
\begin{figure}[h!]
	\centering
     \subfigure[human]
    {\label{fig:instant1:human}\includegraphics[scale=0.37]{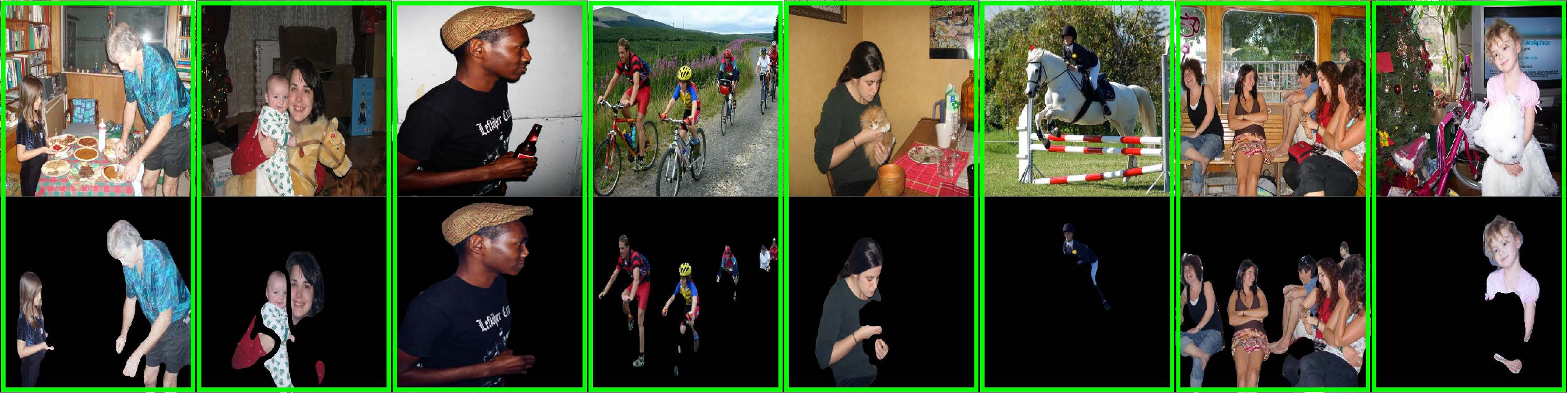}}
    \subfigure[car]
    {\label{fig:instant1:car}\includegraphics[scale=0.39]{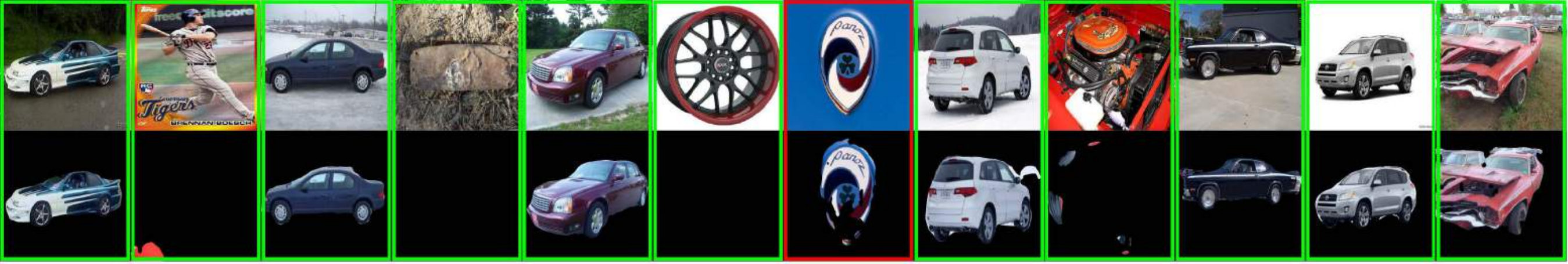}}
    \subfigure[giraffe(unseen)]
    {\label{fig:instant1:giraffe}\includegraphics[scale=0.36]{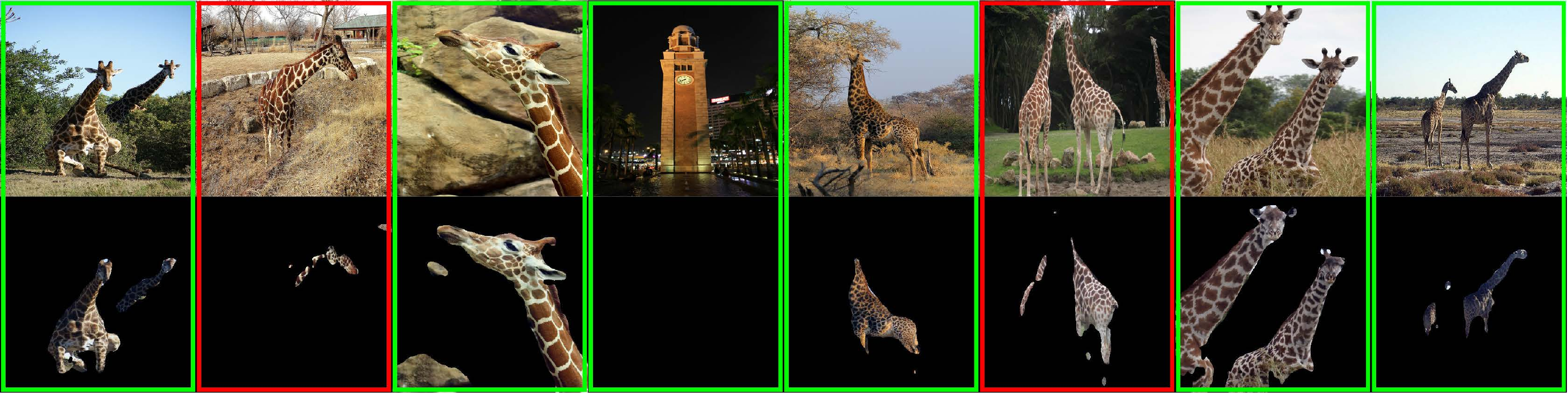}}
	\caption{Results of instant group co-segmentation. The green bounding boxes present the correct answers and red bounding boxes show failures.}
	\label{fig:instant1}
\end{figure}

Algorithm ~\ref{alg::instant} presents each step in instant group co-segmentation. By using this method, we can reduce the time complexity of instant group co-segmentation to linear time complexity. All of the previous work has quadratic time complexity which made it computationally intractable to test on the whole Internet dataset. In Table~\ref{table:instant:Internet}, we show the results using instant group co-segmentation in the whole Internet dataset which contains 1306 car, 879 horse, and 561 airplane images labeled. We reach state-of-the-art performance without doing co-segmentation pair-wisely. Figure~\ref{fig:instant1} shows some qualitative results of our instant group co-segmentation.

\setlength{\tabcolsep}{4pt}
\begin{table}[h!]
\centering
\caption{Instant group co-segmentation results on whole Internet Dataset(time cost 590s). We cannot compare the result with \cite{objectcoseg2} since it's computational intractable.}
\begin{tabular}{cccc}
\hline\noalign{\smallskip}
Avg Jaccard &\cite{objectcoseg4}&\cite{objectcoseg8} & CA-instant \\
\noalign{\smallskip}
\hline
\noalign{\smallskip}
Car  &63.4 &64.7 & \textbf{83.4} \\
Horse  &53.9&57.6& \textbf{70.9} \\
Airplane &55.6&60.0& \textbf{60.6} \\
Average &57.6&60.7& \textbf{71.7}\\
\hline
\end{tabular}
\label{table:instant:Internet}
\end{table}
\setlength{\tabcolsep}{1.4pt}

\setlength{\tabcolsep}{4pt}
\begin{table}[h!]
\centering
\caption{Instant group co-segmentation results on different datasets, compared with results using pairwise co-segmentation.}
\begin{tabular}{ccccc}
\hline\noalign{\smallskip}
Avg Jaccard (test time) & \cite{objectcoseg2}& CA& CA-instant\\
\noalign{\smallskip}
\hline
\noalign{\smallskip}
MSRC(seen)  &\textbf{79.9} (203)& 76.5 (51)& 73.9 (17)\\
Internet(seen) &70.3 (9531)& \textbf{72.8} (2179)& 70.9 (63)\\
ICoseg(unseen) &84.2 (1077)& 85.2 (268)  &\textbf{87.1} (43) \\
\hline
\end{tabular}
\label{table:instant:all}
\end{table}
\setlength{\tabcolsep}{1.4pt}

To show that our instant group co-segmentation method does not sacrifice accuracy compared with previous methods based on pairwise co-segmentation, we carry out the same quantitative experiment in the same dataset as done in Section \ref{sec:resultCoseg}. From Table~\ref{table:instant:all}, we can see that the instant group co-segmentation achieves higher accuracy for \textit{Unseen Objects} of iCoseg dataset, while performing comparably well in \textit{Seen Objects} from MSRC dataset and Internet dataset, compared with \cite{objectcoseg2} and our CA model used pair-wisely. This demonstrates that the global average attention helps us filtering some semantic noises in the irrelevant channels.

There are three typical failure cases in co-segmentation task: loss of segmentation accuracy, over-segmentation, and under-segmentation. For example in Fig.~\ref{fig:instant1:car}, the 7th image is over-segmented and in Fig.~\ref{fig:instant1:giraffe}, the 2nd image is under-segmented.
Over-segment and under-segment can be easily avoided in \textit{Seen Objects} as long as the model converges.
However, \textit{Unseen Objects} are beyond control during training.

According to \cite{pixelobjectness}, the pretrained model is well capable of segmenting unseen objects. 
Our models find the balance between attention learners and the pretrained model during training. If the pretrained model dominates, it tends to segment all objects in the image, resulting in over-segment.
On the other hand, if the attention learner overfits, unseen objects would be ignored, leading to under-segment.

\begin{figure}[h!]
    \includegraphics[width=\linewidth]{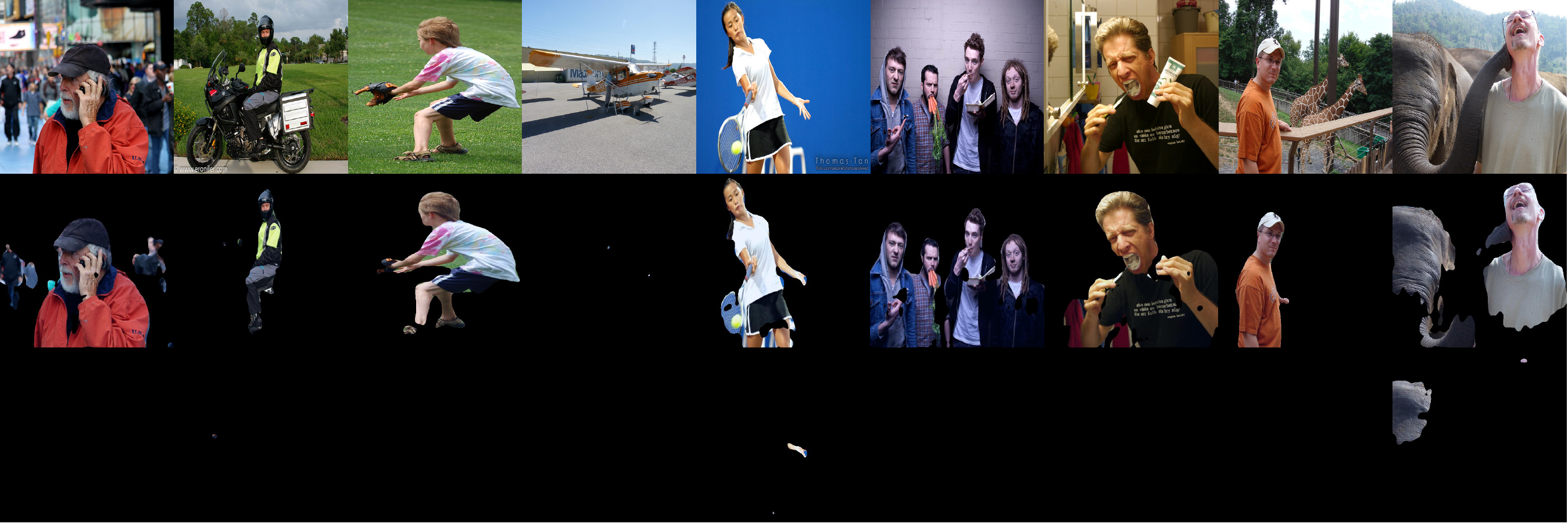}
	\caption{Compare segmentation results with two processions. The 2nd row shows the results using the \textbf{Global Average Attention}: 8/9 of the images contain human class, so humans will be segmented in this case. The 3rd row demonstrates the results using the \textbf{Global Minimum Attention}. In the 4th image, there is no human, so nothing will be segmented in all images with this method.}
	\label{fig:instant3}
\end{figure}

\textbf{Group Average Attention} in Section \ref{sec:group} is not the only meaningful procession in our model. For other tasks, such as segmenting the common objects that exist in all images, we can choose the minimum weight for each channel from the generated attention and obtain \textbf{Global Minimum Attention}.
Fig.~\ref{fig:instant3} shows an example of comparing global average attention and global minimum attention. \textbf{Group Average Attention} can segment the most common object in multiple images, while \textbf{Group Minimum Attention} strictly finds the objects appear in all images.

\section{Conclusion}
In this paper, we propose three different architectures of attention learner as semantic selectors. With proposed instant group co-segmentation, we can co-segment multiple images in the linear time. We present that our results have achieved state-of-the-art in the object co-segmentation task and outperformed \cite{pixelobjectness} in object discovering ability. We also visualize the channel-wise attention and the spatial-wise attention to show the correctness of our proposed models.

However, using our proposed model, if the inputs do not contain the same object, our model will output wrong results since there's no common semantic class. Together with applying zero-shot learning to improve segmentation of unseen objects, we leave this for our future work.

\section{Appendix}
\subsection{Visualization of generated attention}
To make sure that the channel wise attention has obeyed our assumption that enhance specific channels and suppress others, we generate channel wise attention $\bm{\alpha} \in R^{512}$ from five different classes: \emph{Monitor}, \emph{Indigo bird}, \emph{Sheep}, \emph{Jellyfish}, \emph{Shopping Cart}. Here, each classes contain 1300 images from Imagenet \cite{imagenet}. 

\begin{figure}[!htb]
	\centering
	\includegraphics[height=4cm]{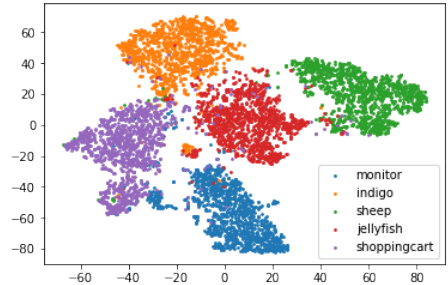}
	\caption{Visualization of generated attention with t-SNE \cite{tsne}. We can see that different classes generated different attention. They are disentangled.}
	\label{fig:tsne}
\end{figure}

Fig.~\ref{fig:tsne} shows the results. We can see that images in the same classes generate the similar attention. Those images in different classes generate different attention. So it proves that channel wise attention enhances some channels and suppress some channels according to the classes in images.

\subsection{Visualization of spatial attention}

\begin{figure}
	\centering
	\includegraphics[scale=0.45]{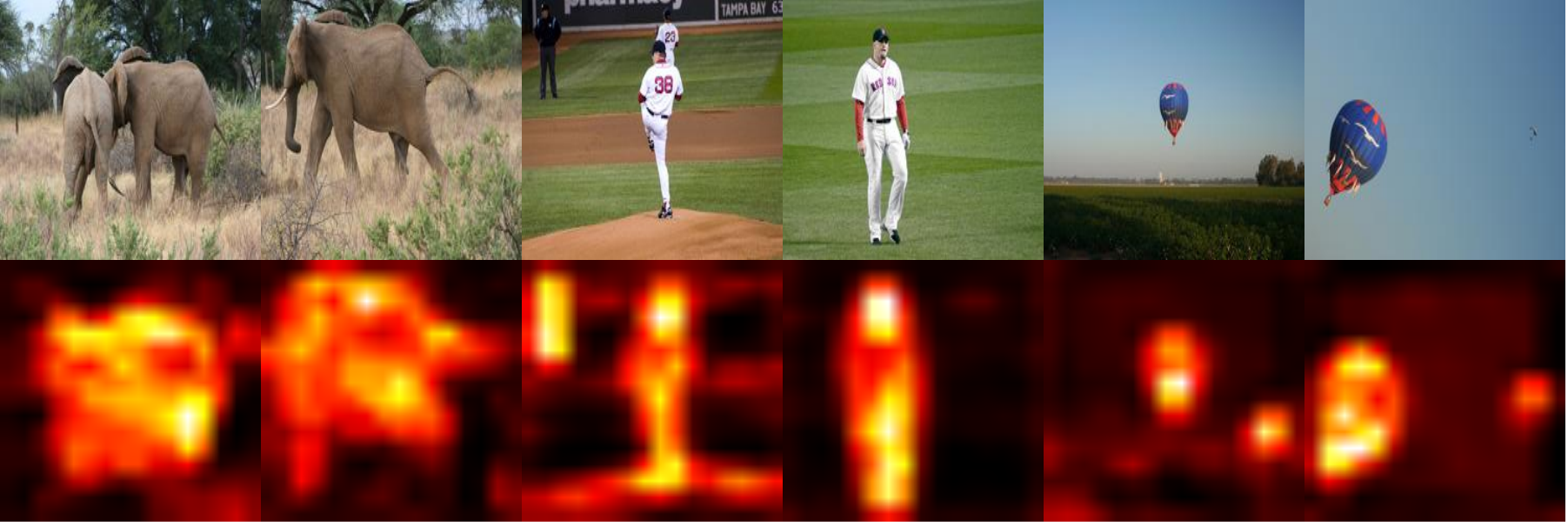}
	\caption{Visualization of the spatial-wise attention. Red indicates high values, helping feature map to recover the semantic information lost by global average pooling. Here the spatial-wise attention is generated before semantic selector, so it will pay "attention" to each object-like pixel.}
	\label{fig:heat}
\end{figure}

In CSA, we improve our model with an additional spatial attention. After Global Average Pooling (GAP), some spatial information has been lost. Thus we use spatial attention to enhance some important spatial locations in the images.
Fig.\ref{fig:heat} shows the visualization of generated spatial attention map.

\vspace{3cm}

\bibliographystyle{splncs04}
\bibliography{accv2018cameraready}

\end{document}